%% file: main.tex
\documentclass{article} 
\usepackage{iclr2025_conference,times}
\usepackage{graphicx}

\input{math_commands.tex}

\input{defs}
\usepackage[colorlinks=true, linkcolor=blue, citecolor=blue, urlcolor=blue]{hyperref}
\usepackage{url}
\usepackage{lipsum}
\usepackage{xspace}
\usepackage{subfigure}

\usepackage{graphicx}
\usepackage{booktabs}
\usepackage{algorithm}
\usepackage{algpseudocode}
\usepackage{listings}
\usepackage{upquote}  
\usepackage{xcolor,colortbl}
\usepackage{enumitem}
\lstdefinestyle{overleaf}{
    backgroundcolor=\color[rgb]{0.95,0.95,0.92},   
    commentstyle=\color[rgb]{0,0.6,0},
    keywordstyle=\color{magenta},
    numberstyle=\tiny\color[rgb]{0.5,0.5,0.5},
    stringstyle=\color[rgb]{0.58,0,0.82},
    basicstyle=\ttfamily\footnotesize,
    breakatwhitespace=false,         
    breaklines=true,                 
    captionpos=b,                    
    keepspaces=true,                 
    numbers=left,                    
    numbersep=5pt,                  
    showspaces=false,                
    showstringspaces=false,
    showtabs=false,                  
    tabsize=2
}
\lstdefinestyle{mocov3}{
  backgroundcolor=\color{white},
  basicstyle=\fontsize{7.5pt}{7.5pt}\ttfamily\selectfont,
  columns=fullflexible,
  breaklines=true,
  captionpos=b,
  commentstyle=\fontsize{7.5pt}{7.5pt}\color[rgb]{0.25,0.5,0.5},
  keywordstyle=\fontsize{7.5pt}{7.5pt}\color[rgb]{0.85,0.18,0.50},
}
\usepackage{amssymb}

\usepackage{tcolorbox}
\tcbuselibrary{skins}

\newtcolorbox{promptbox}{
    enhanced,
    colback=gray!5,
    colframe=gray!40,
    boxrule=0.5pt,
    arc=3mm,
    top=2mm,
    bottom=2mm,
    left=3mm,
    right=3mm,
    width=\textwidth,
    halign=left,
    nobreak,    
    before=\vspace{2mm},
    after=\vspace{2mm}
}
\newtcolorbox{figurebox}{
    enhanced,
    colback=gray!5,
    colframe=gray!40,
    boxrule=0.5pt,
    arc=3mm,
    width=\textwidth,
    before=\vspace{2mm},
    after=\vspace{2mm}
}

\title{\fontsize{15}{19}\selectfont EP-CFG: Energy-Preserving Classifier-Free Guidance}

\author{
    Kai Zhang \hspace{2.0cm} Fujun Luan \hspace{2.0cm} Sai Bi \hspace{2.0cm} Jianming Zhang \\[2mm]
    \centerline{{\bf Adobe Foundation Model Team}} 
}

\iclrfinalcopy 
\begin{document}

\maketitle

\input{sections/ep_cfg}
\input{sections/results}

\clearpage
\newpage

\bibliography{iclr2025_conference}
\bibliographystyle{iclr2025_conference}


\end{document}

%% file: math_commands.tex
\usepackage{amsmath,amsfonts,bm}

\def\eqref#1{equation~\ref{#1}}

\def\1{\bm{1}}

\DeclareMathAlphabet{\mathsfit}{\encodingdefault}{\sfdefault}{m}{sl}
\SetMathAlphabet{\mathsfit}{bold}{\encodingdefault}{\sfdefault}{bx}{n}



%% file: sections/ep_cfg.tex

Classifier-free guidance (CFG)~\citep{ho2022classifier} is widely used in diffusion models~\citep{ho2020denoising, song2020score} for text-guided generation, but often leads to over-contrast and over-saturation artifacts. We propose EP-CFG, a simple yet effective CFG solution that preserves the energy distribution of the conditional prediction while maintaining strong semantic alignment.

Concretely, given unconditional prediction $\mathbf{x}_u$ and conditional prediction $\mathbf{x}_c$, CFG~\citep{ho2022classifier} produces:
\begin{equation}
    \mathbf{x}_{\text{cfg}} = \mathbf{x}_c + (\lambda-1)(\mathbf{x}_c - \mathbf{x}_u), \label{eq:cfg-rewrite}
\end{equation}
where $\lambda \geq 1$ is the CFG strength. Here, we re-write the CFG equation from its common form $    \mathbf{x}_{\text{cfg}} = \lambda \mathbf{x}_c + (1-\lambda)\mathbf{x}_u$~\citep{ho2022classifier} to Eq.~\ref{eq:cfg-rewrite}, in order to make it clearer to see the two components of CFG prediction $\mathbf{x}_{\text{cfg}}$: 1) conditional prediction $\mathbf{x}_c$; 2) CFG additive term $(\lambda-1)(\mathbf{x}_c - \mathbf{x}_u)$. 

The CFG additive term $(\lambda-1)(\mathbf{x}_c - \mathbf{x}_u)$ added to the conditional term $\mathbf{x}_c$ can significantly 
enhance the semantic alignment and structure coherence. Usually, the CFG strength is around 7-10~\citep{rombach2022high} in modern text-to-image models for sampling high-quality visuals. However, it is well-known that such high CFG strength can lead to the well-known over-contrast and over-saturation artifacts~\citep{ho2022classifier}. Concurrent work~\citep{sadat2024eliminating} proposed APG to address oversaturation through update term decomposition. Our key observation is that the CFG additive component can drastically increase the energy of the sampled latents in the progressive denoising process; the excessive energy eventually causes the over-contrast and over-saturation artifacts.


As a remedy, our proposed EP-CFG rescales the energy of $\mathbf{x}_{\text{cfg}}$ to match $\mathbf{x}_c$ at each denoising step:
\begin{equation}
    \mathbf{x}'_{\text{cfg}} = \mathbf{x}_{\text{cfg}} \cdot \sqrt{\frac{E_c}{E_{\text{cfg}}}},
\end{equation}
where $E_c = \|\mathbf{x}_c\|^2$ and $E_{\text{cfg}} = \|\mathbf{x}_{\text{cfg}}\|^2$. Our EP-CFG differs from prior solution~\citep{lin2024common} mainly in two aspects: 
1) we scale the CFG prediction based on energy instead of standard deviation; 2) we do not interpolate between CFG and conditional predictions. Hence, our formulation is simpler and better preserves the information of the CFG prediction.

To increase our algorithm's robustness, we ignore the two tails in the energy histogram and only consider a small middle region when estimating the energy terms $E_c,E_{\text{cfg}}$. Specifically, let $P_l$, $P_h$ be the $l^{\text{th}}$ and $h^{\text{th}}$ percentiles of a latent $\mathbf{x}$; then the robust energy is derived as:
\begin{equation}
E^{\text{robust}} = \sum_{i} \mathbf{x}_i^2 \cdot \mathbf{1}[P_l \leq \mathbf{x}_i^2 \leq P_h].
\end{equation}
In practice, we recommend using $l=45$, $h=55$ as we found this setting works best empirically, and all results in this report use these values. By using the robust energy estimation, we suppress the confetti artifacts that oftentimes appear in monochromatic regions of generated visuals.

%% file: sections/results.tex
\begin{figure}
\vspace*{-1.6cm}
\begin{figurebox}
    \textbf{Prompt:} \textit{An astronaut running through an alley in Rio de Janeiro.}
    \vspace{2mm}
    
    \centering
    \subfigure[CFG level=5, EP-CFG OFF]{
        \includegraphics[width=0.48\textwidth]{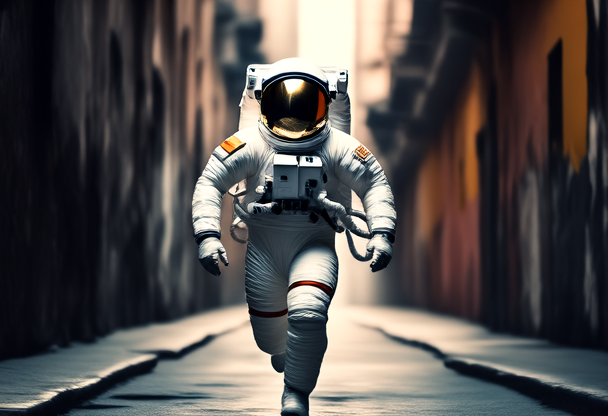}
    }
    \subfigure[CFG level=5, EP-CFG ON]{
        \includegraphics[width=0.48\textwidth]{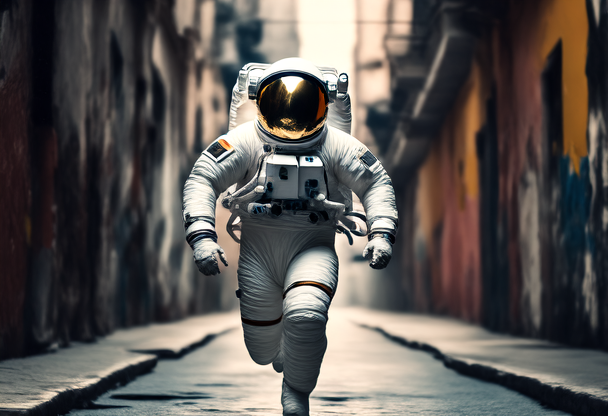}
    }
    
    \subfigure[CFG level=7, EP-CFG OFF]{
        \includegraphics[width=0.48\textwidth]{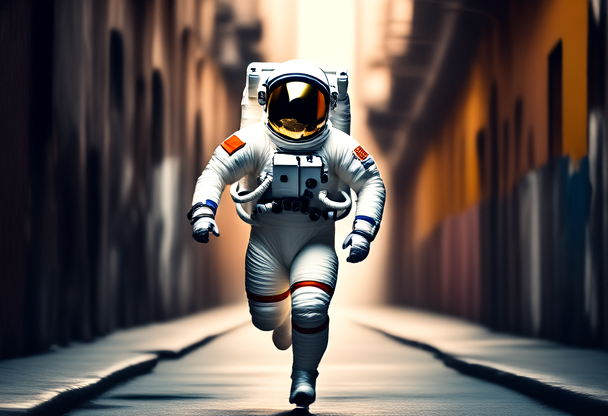}
    }
    \subfigure[CFG level=7, EP-CFG ON]{
        \includegraphics[width=0.48\textwidth]{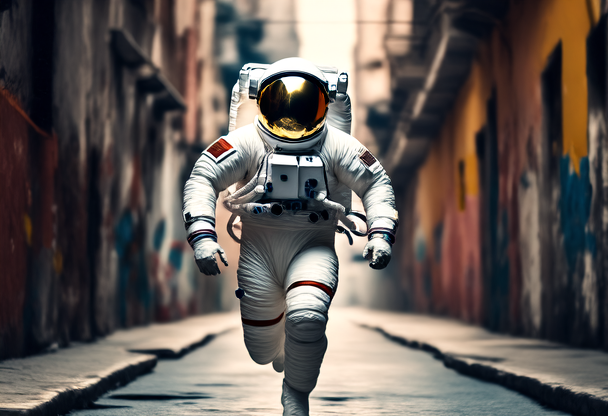}
    }
    
    \subfigure[CFG level=9, EP-CFG OFF]{
        \includegraphics[width=0.48\textwidth]{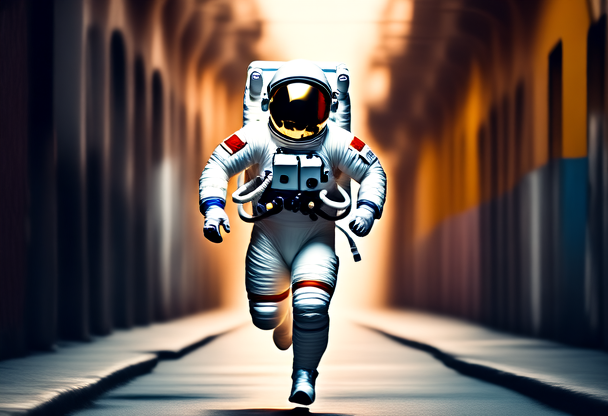}
    }
    \subfigure[CFG level=9, EP-CFG ON]{
        \includegraphics[width=0.48\textwidth]{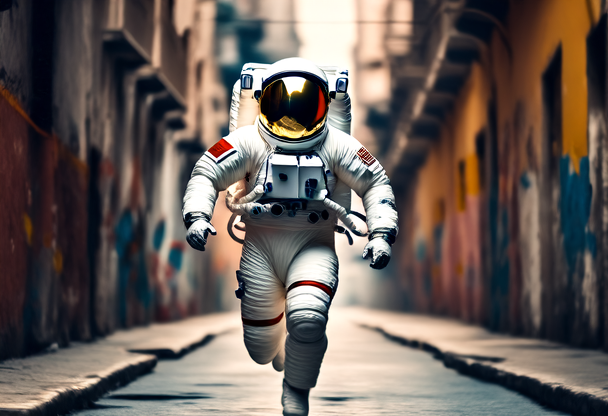}
    }
    
    \subfigure[CFG level=12, EP-CFG OFF]{
        \includegraphics[width=0.48\textwidth]{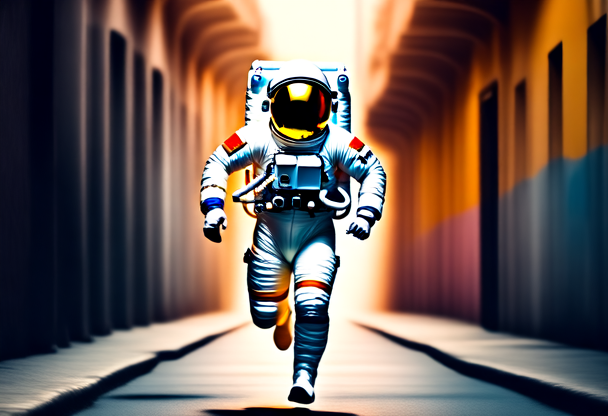}
    }
    \subfigure[CFG level=12, EP-CFG ON]{
        \includegraphics[width=0.48\textwidth]{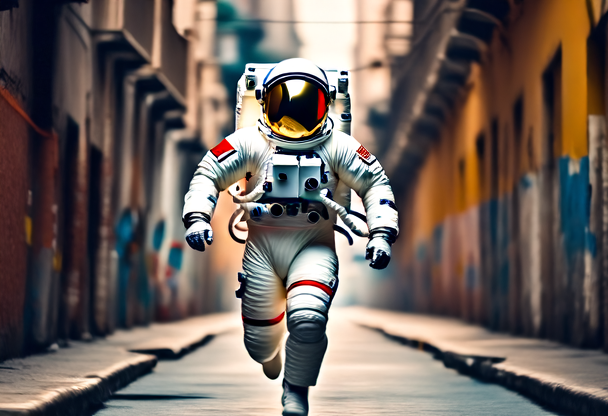}
    }
   
    \caption{Comparison of generation results with different CFG strengths (increasing from top to bottom). Without EP-CFG (left column), higher CFG strengths lead to over-saturation and contrast artifacts, especially in the background. EP-CFG (right column) maintains natural image quality across all CFG strengths while preserving the benefits of stronger guidance.}
    
    \label{fig:filix_cfg_comparison1}
    \end{figurebox}
\end{figure}

\begin{figure}
\vspace*{-1.6cm}
\begin{figurebox}
    \textbf{Prompt:} \textit{Motorcycle running insanely fast through an enchanted forrest. It enters a cave, inside it is New York City. The camera follows it while it drifts and runs. Photorealistic style.}
    
    \centering
    \subfigure[CFG level=5, EP-CFG OFF]{
        \includegraphics[width=0.48\textwidth]{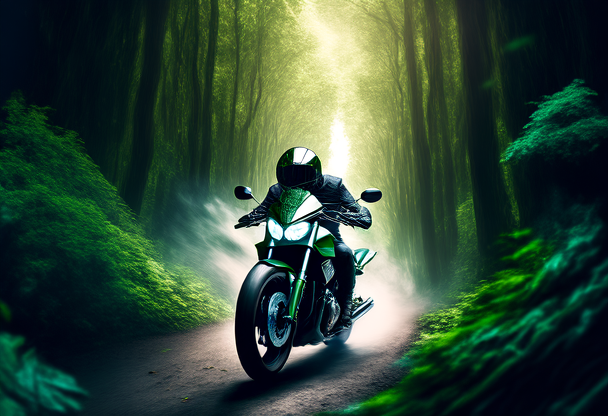}
    }
    \subfigure[CFG level=5, EP-CFG ON]{
        \includegraphics[width=0.48\textwidth]{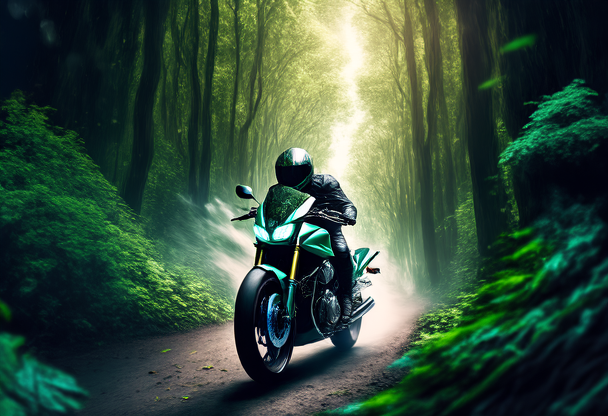}
    }
    
    \subfigure[CFG level=7, EP-CFG OFF]{
        \includegraphics[width=0.48\textwidth]{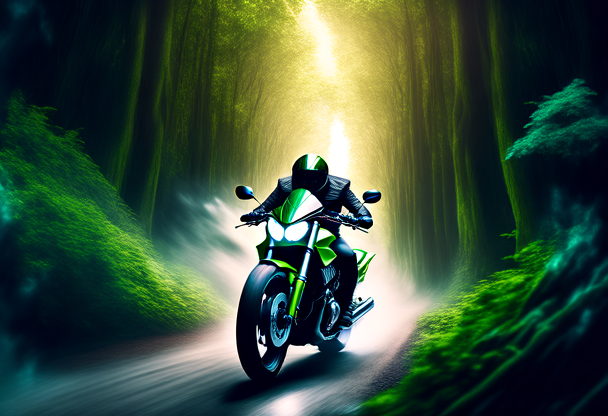}
    }
    \subfigure[CFG level=7, EP-CFG ON]{
        \includegraphics[width=0.48\textwidth]{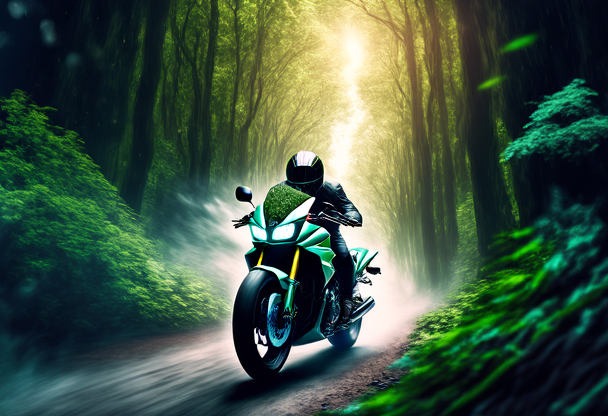}
    }
    
    \subfigure[CFG level=9, EP-CFG OFF]{
        \includegraphics[width=0.48\textwidth]{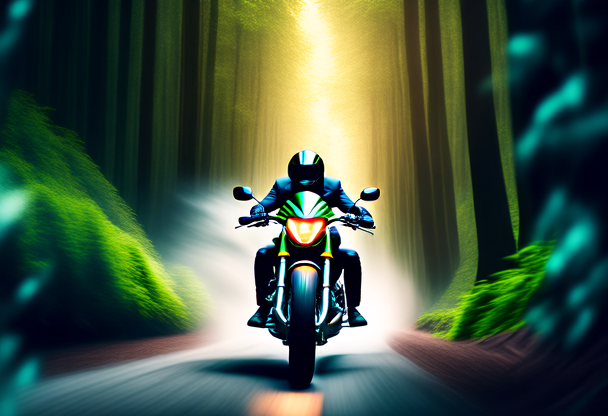}
    }
    \subfigure[CFG level=9, EP-CFG ON]{
        \includegraphics[width=0.48\textwidth]{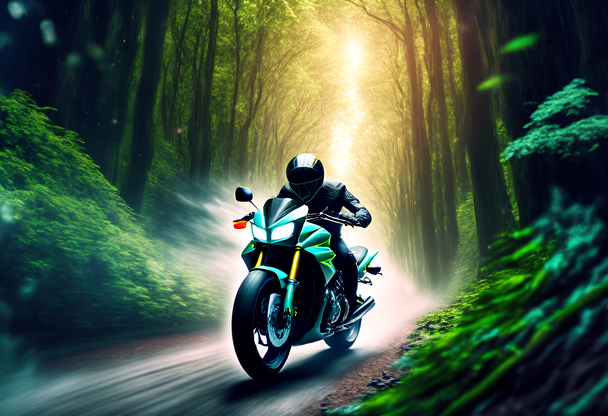}
    }
    
    \subfigure[CFG level=12, EP-CFG OFF]{
        \includegraphics[width=0.48\textwidth]{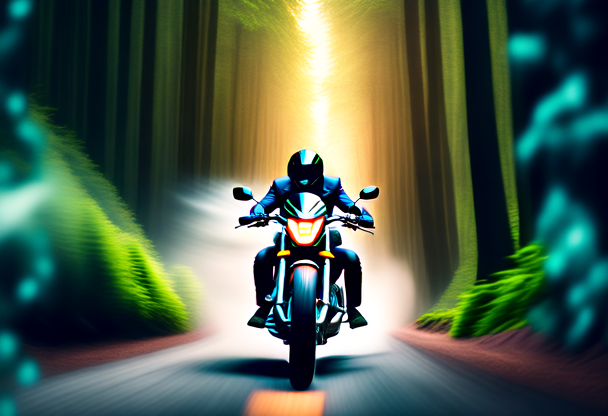}
    }
    \subfigure[CFG level=12, EP-CFG ON]{
        \includegraphics[width=0.48\textwidth]{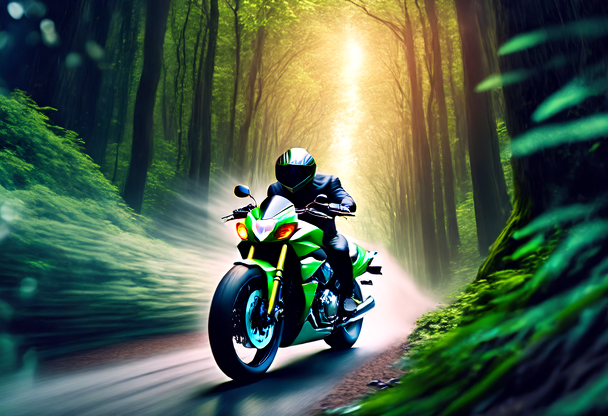}
    }
   
    \caption{Comparison of generation results with different CFG strengths. Without EP-CFG (left column), higher CFG strengths cause severe over-contrast and light bloom effects in the forest scene. EP-CFG (right column) preserves natural lighting and details across all guidance scales while maintaining the motorcycle's structure and forest atmosphere.}
    
    \label{fig:filix_cfg_comparison2}
    \end{figurebox}
\end{figure}

\begin{figure}
\vspace*{-1.6cm}
\begin{figurebox}
    \textbf{Prompt:} \textit{A Chinese Lunar New Year celebration video with Chinese Dragon.}
    
    \centering
    \subfigure[CFG level=5, EP-CFG OFF]{
        \includegraphics[width=0.48\textwidth]{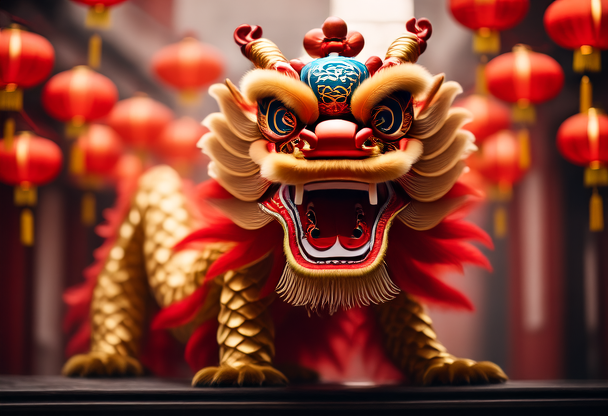}
    }
    \subfigure[CFG level=5, EP-CFG ON]{
        \includegraphics[width=0.48\textwidth]{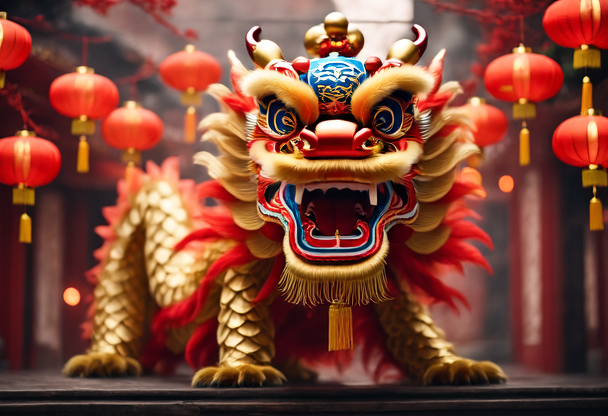}
    }
    
    \subfigure[CFG level=7, EP-CFG OFF]{
        \includegraphics[width=0.48\textwidth]{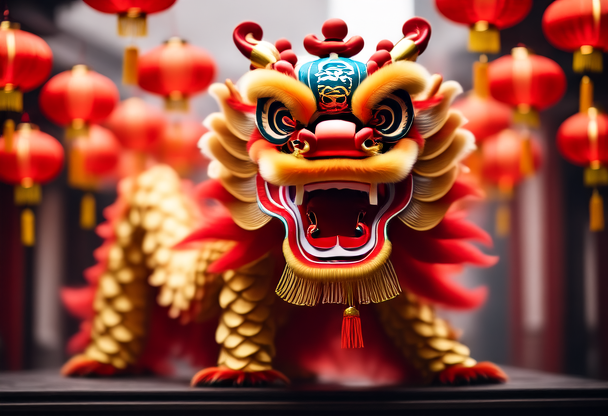}
    }
    \subfigure[CFG level=7, EP-CFG ON]{
        \includegraphics[width=0.48\textwidth]{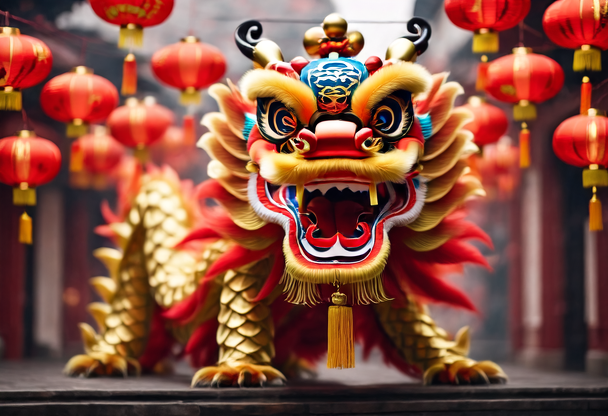}
    }
    
    \subfigure[CFG level=9, EP-CFG OFF]{
        \includegraphics[width=0.48\textwidth]{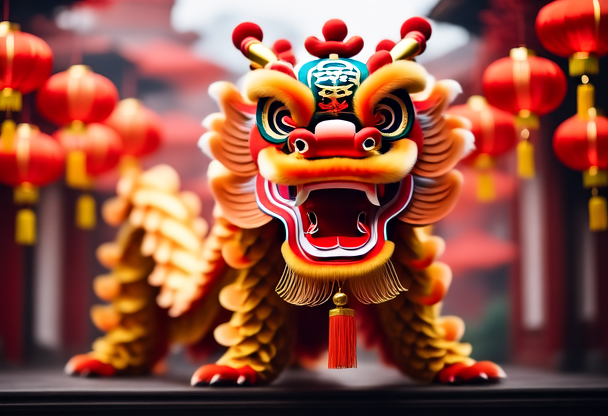}
    }
    \subfigure[CFG level=9, EP-CFG ON]{
        \includegraphics[width=0.48\textwidth]{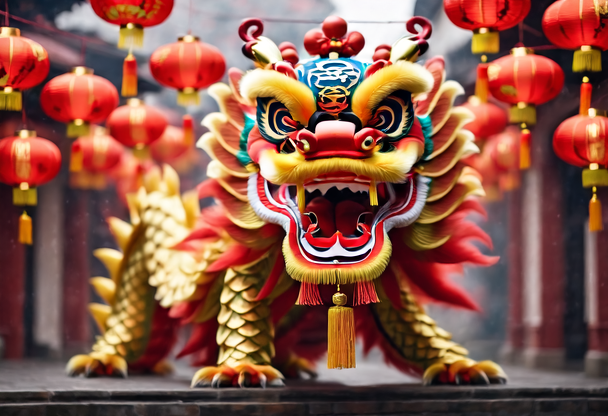}
    }
    
    \subfigure[CFG level=12, EP-CFG OFF]{
        \includegraphics[width=0.48\textwidth]{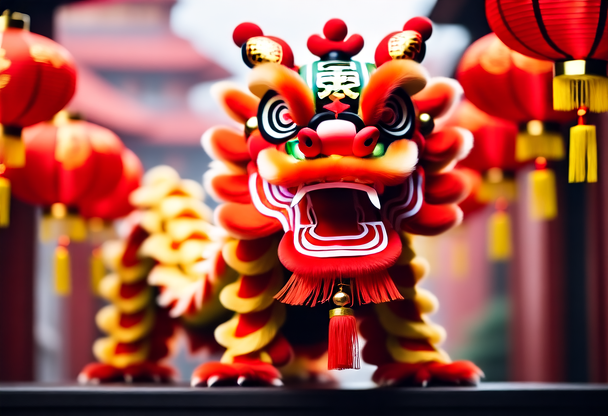}
    }
    \subfigure[CFG level=12, EP-CFG ON]{
        \includegraphics[width=0.48\textwidth]{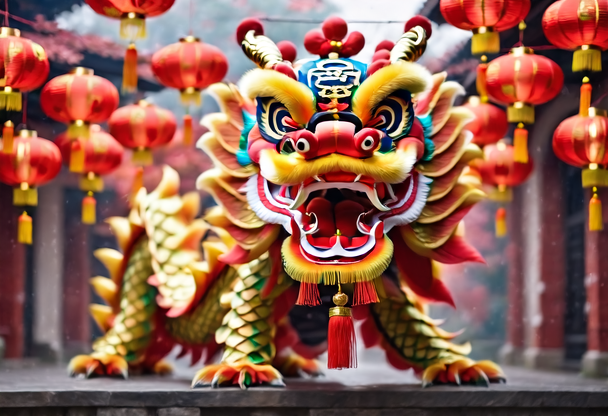}
    }
   
    \caption{Comparison of generation results with different CFG strengths (increasing from top to bottom). Without EP-CFG (left column), higher CFG strengths lead to over-contrast and color saturation in the dragon's features and lantern decorations. EP-CFG (right column) maintains natural image quality and preserves intricate details of the dragon's ornate design across all CFG strengths while retaining the festive Lunar New Year ambiance.}
    
    \label{fig:filix_cfg_comparison3}
    \end{figurebox}
\end{figure}

\begin{figure}
\vspace*{-1.6cm}
\begin{figurebox}
    \textbf{Prompt:} \textit{A flower growing out of the windowsill of a suburban house.}
    
    \centering
    \subfigure[CFG level=5, EP-CFG OFF]{
        \includegraphics[width=0.48\textwidth]{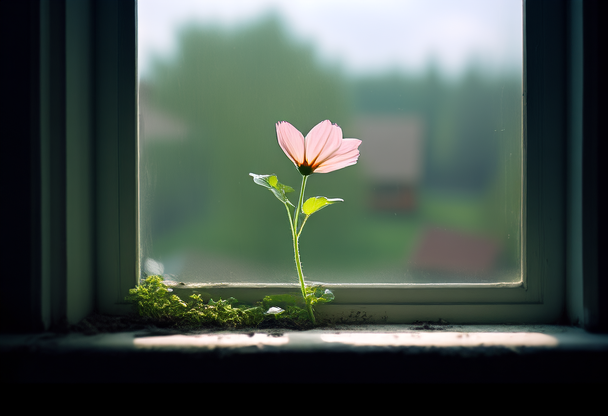}
    }
    \subfigure[CFG level=5, EP-CFG ON]{
        \includegraphics[width=0.48\textwidth]{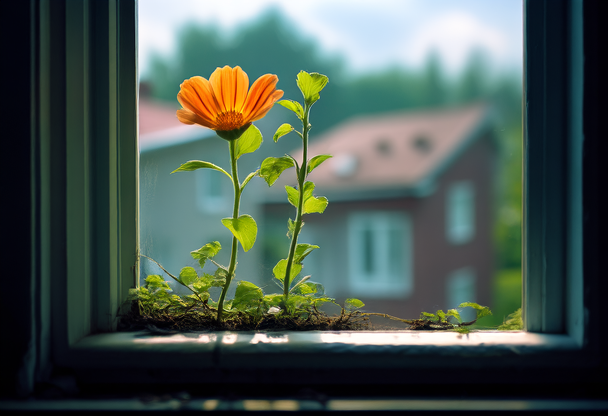}
    }
    
    \subfigure[CFG level=7, EP-CFG OFF]{
        \includegraphics[width=0.48\textwidth]{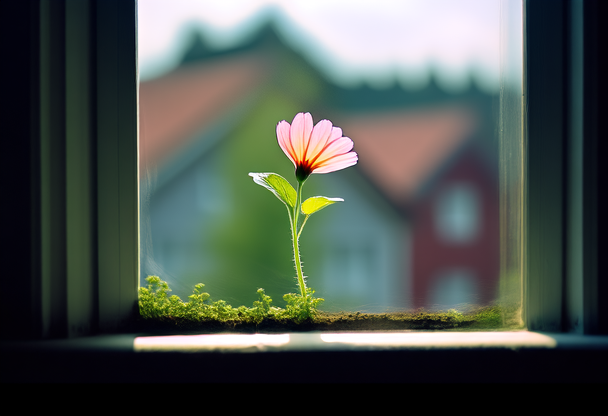}
    }
    \subfigure[CFG level=7, EP-CFG ON]{
        \includegraphics[width=0.48\textwidth]{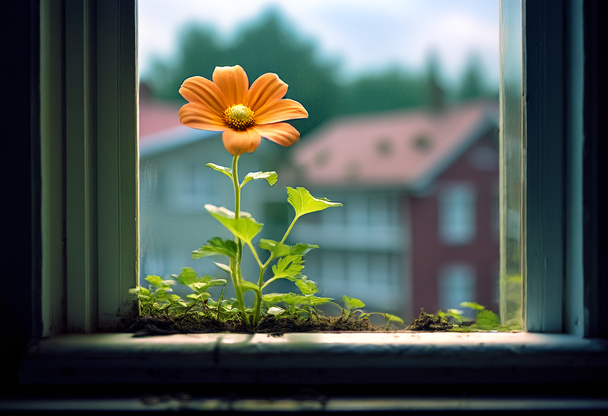}
    }
    
    \subfigure[CFG level=9, EP-CFG OFF]{
        \includegraphics[width=0.48\textwidth]{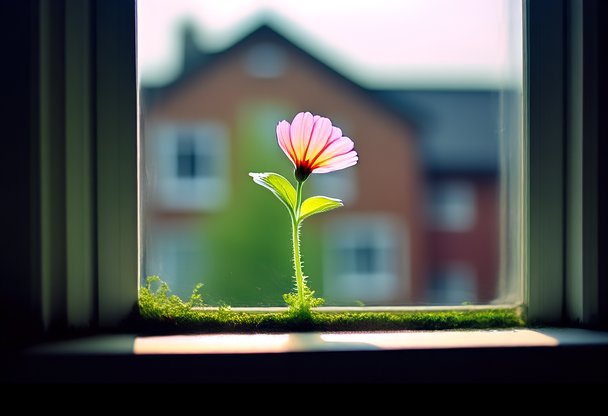}
    }
    \subfigure[CFG level=9, EP-CFG ON]{
        \includegraphics[width=0.48\textwidth]{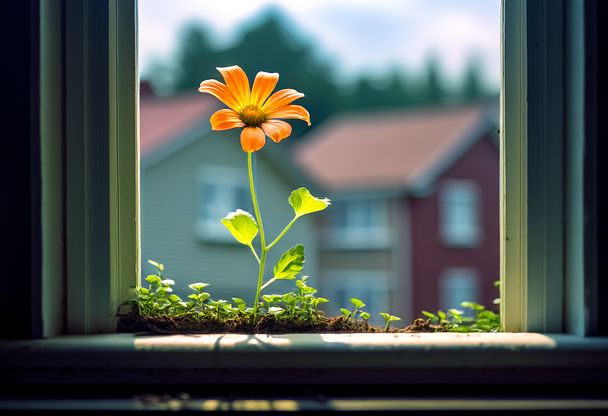}
    }
    
    \subfigure[CFG level=12, EP-CFG OFF]{
        \includegraphics[width=0.48\textwidth]{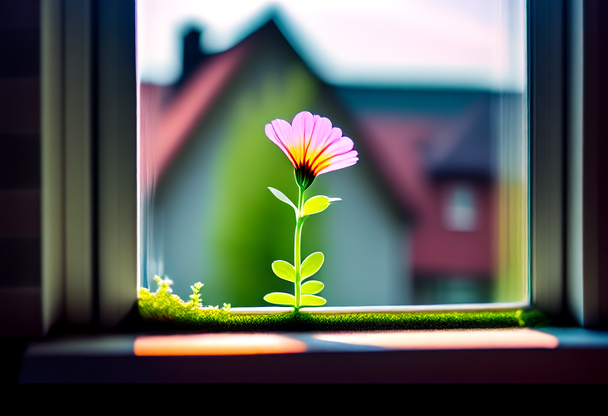}
    }
    \subfigure[CFG level=12, EP-CFG ON]{
        \includegraphics[width=0.48\textwidth]{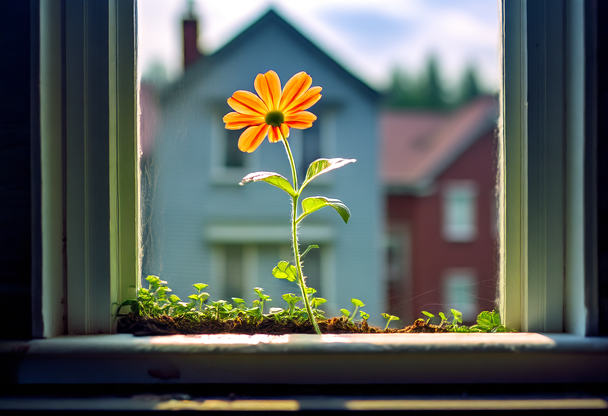}
    }
   
    \caption{Comparison of generation results with different CFG strengths (increasing from top to bottom). Without EP-CFG (left column), higher CFG strengths lead to over-contrast and color saturation in the flower petals and suburban background. EP-CFG (right column) maintains natural image quality and preserves delicate details of the flower and foliage across all CFG strengths while retaining the serene windowsill setting.}
    \label{fig:filix_cfg_comparison4}
    \end{figurebox}
\end{figure}

\begin{figure}
\vspace*{-1.6cm}
\begin{figurebox}
    \textbf{Prompt:} \textit{A horse galloping through van Gogh's Starry Night.}
    
    \centering
    \subfigure[CFG level=5, EP-CFG OFF]{
        \includegraphics[width=0.48\textwidth]{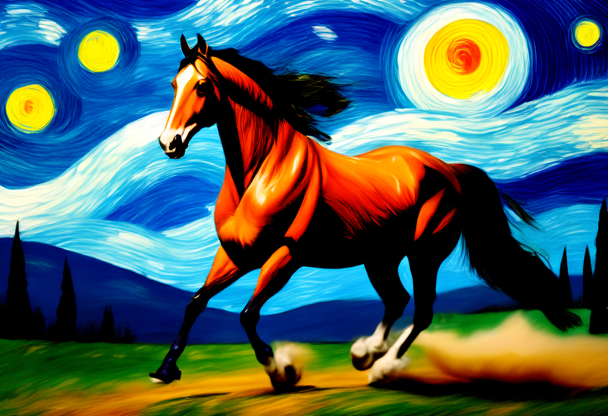}
    }
    \subfigure[CFG level=5, EP-CFG ON]{
        \includegraphics[width=0.48\textwidth]{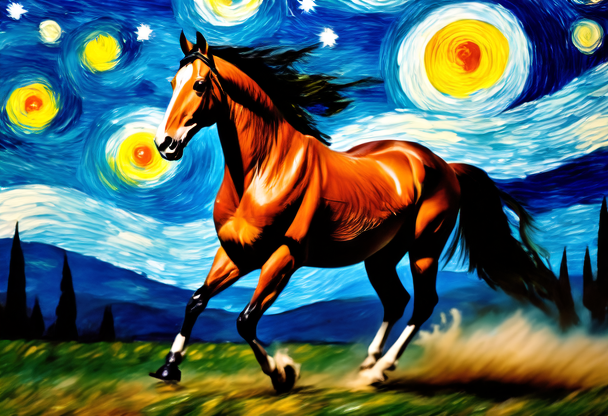}
    }
    
    \subfigure[CFG level=7, EP-CFG OFF]{
        \includegraphics[width=0.48\textwidth]{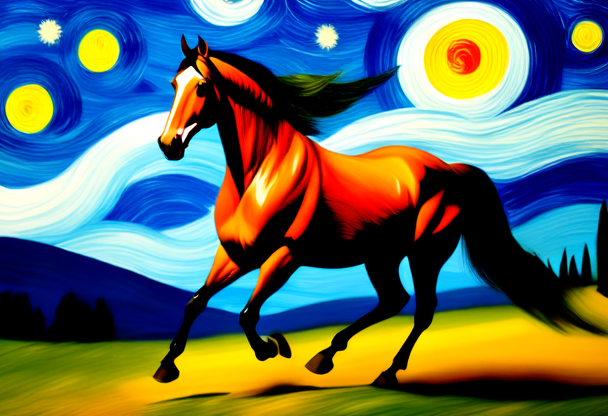}
    }
    \subfigure[CFG level=7, EP-CFG ON]{
        \includegraphics[width=0.48\textwidth]{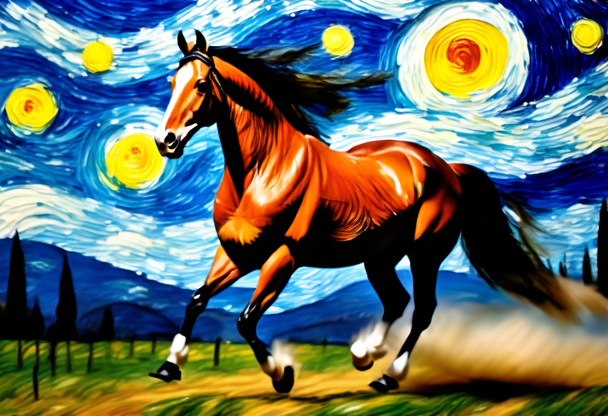}
    }
    
    \subfigure[CFG level=9, EP-CFG OFF]{
        \includegraphics[width=0.48\textwidth]{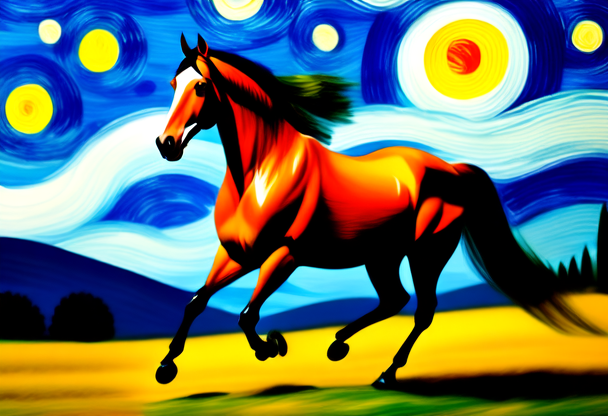}
    }
    \subfigure[CFG level=9, EP-CFG ON]{
        \includegraphics[width=0.48\textwidth]{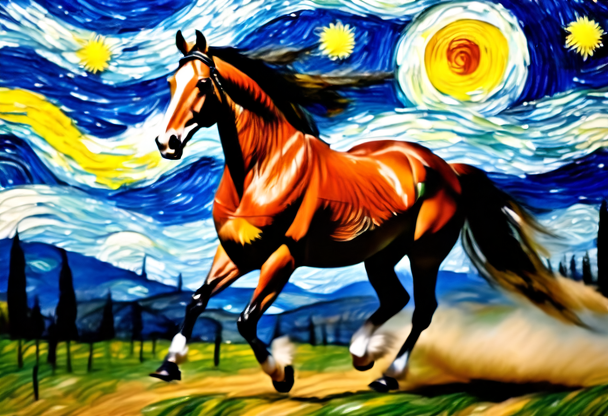}
    }
    
    \subfigure[CFG level=12, EP-CFG OFF]{
        \includegraphics[width=0.48\textwidth]{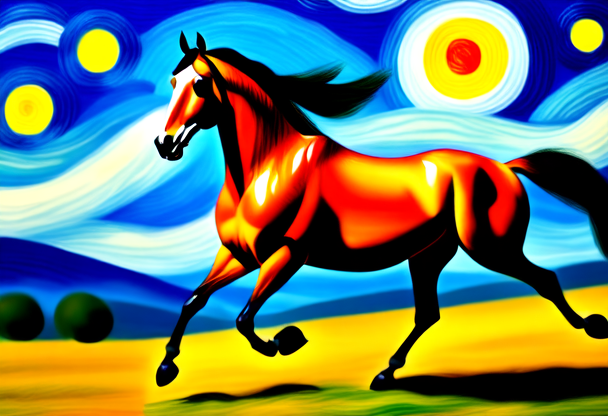}
    }
    \subfigure[CFG level=12, EP-CFG ON]{
        \includegraphics[width=0.48\textwidth]{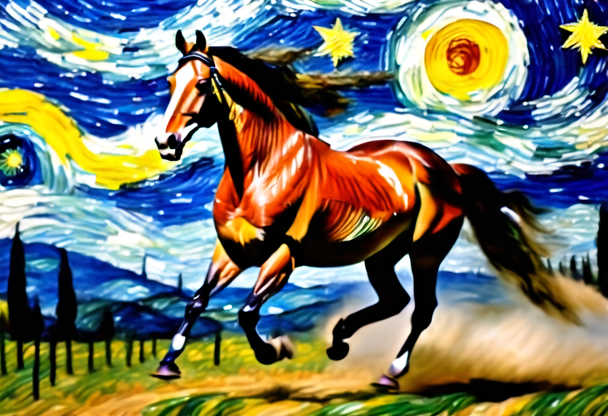}
    }
   
    \caption{Comparison of generation results with different CFG strengths (increasing from top to bottom). Without EP-CFG (left column), higher CFG strengths lead to over-contrast and color distortion in the horse's form and Van Gogh's signature swirling sky. EP-CFG (right column) maintains natural image quality and preserves the characteristic brushstroke details across all CFG strengths while retaining the distinctive Starry Night artistic style.}
    \label{fig:filix_cfg_comparison5}
    \end{figurebox}
\end{figure}

\begin{figure}
\vspace*{-1.6cm}
\begin{figurebox}
    \textbf{Prompt:} \textit{The Orient Express driving through a fantasy landscape, oil on canvas.}
    
    \centering
    \subfigure[CFG level=5, EP-CFG OFF]{
        \includegraphics[width=0.48\textwidth]{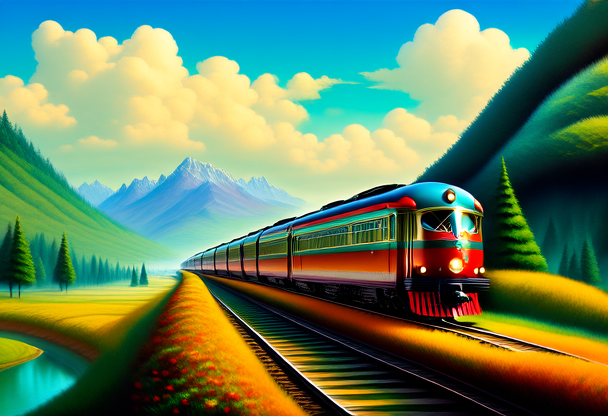}
    }
    \subfigure[CFG level=5, EP-CFG ON]{
        \includegraphics[width=0.48\textwidth]{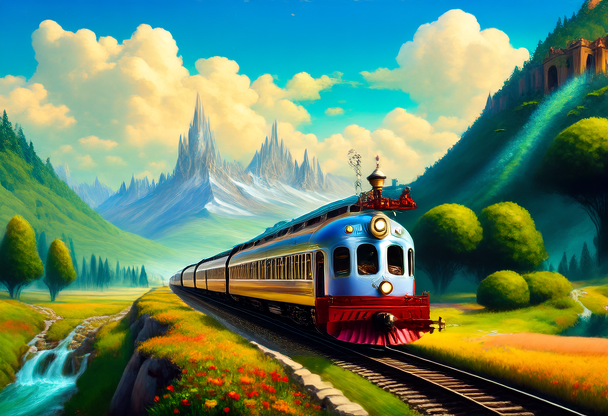}
    }
    
    \subfigure[CFG level=7, EP-CFG OFF]{
        \includegraphics[width=0.48\textwidth]{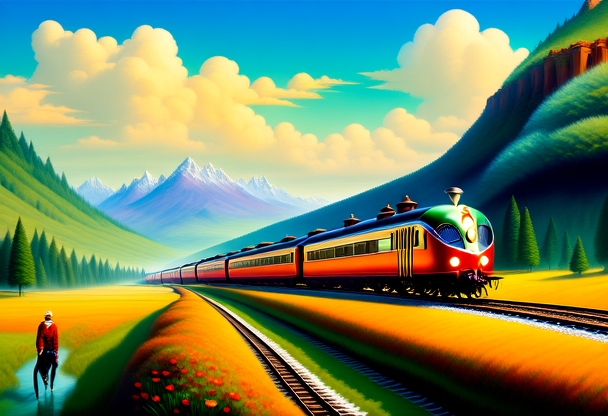}
    }
    \subfigure[CFG level=7, EP-CFG ON]{
        \includegraphics[width=0.48\textwidth]{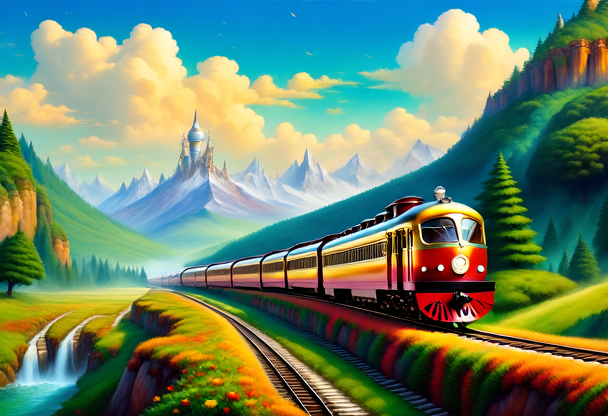}
    }
    
    \subfigure[CFG level=9, EP-CFG OFF]{
        \includegraphics[width=0.48\textwidth]{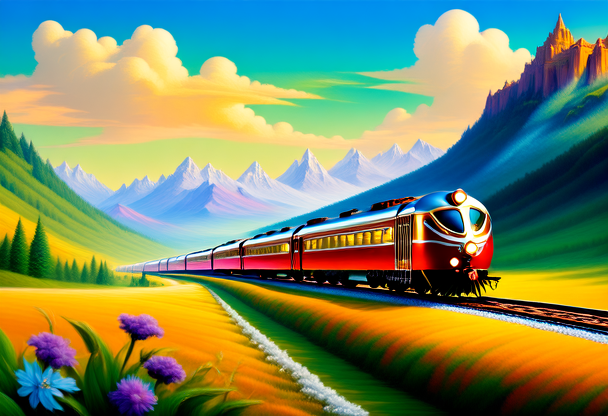}
    }
    \subfigure[CFG level=9, EP-CFG ON]{
        \includegraphics[width=0.48\textwidth]{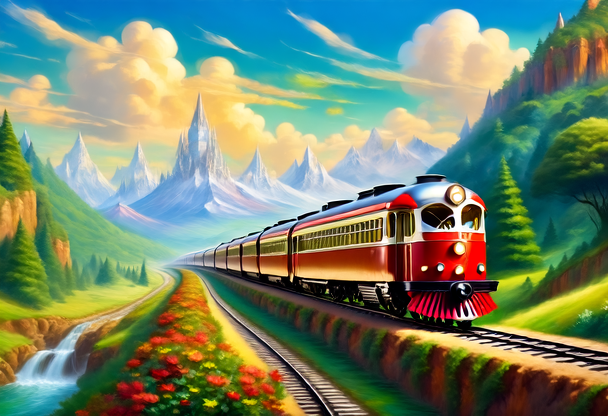}
    }
    
    \subfigure[CFG level=12, EP-CFG OFF]{
        \includegraphics[width=0.48\textwidth]{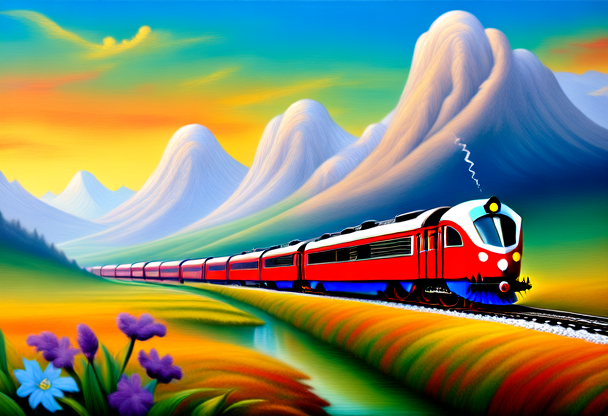}
    }
    \subfigure[CFG level=12, EP-CFG ON]{
        \includegraphics[width=0.48\textwidth]{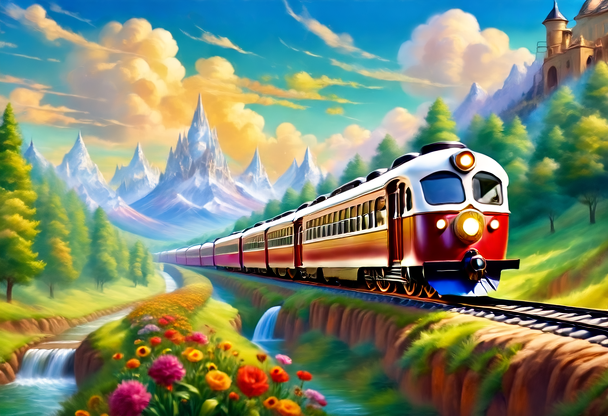}
    }
   
    \caption{Comparison of generation results with different CFG strengths (increasing from top to bottom). Without EP-CFG (left column), higher CFG strengths lead to over-contrast and color saturation in the train's details and fantasy landscape elements. EP-CFG (right column) maintains natural image quality and preserves intricate details of both the Orient Express and the scenic environment across all CFG strengths while retaining the oil painting aesthetics.}
    \label{fig:filix_cfg_comparison6}
    \end{figurebox}
\end{figure}

\begin{figure}
\vspace*{-1.6cm}
\begin{figurebox}
    \textbf{Prompt:} \textit{In a lavish banquet hall adorned with gold and silver dishes, crystal chandeliers casting a warm glow, and silk curtains draped elegantly from the high ceiling, a group of guests dressed in extravagant Versailles, Rococo, and Baroque inspired costumes are engaged in lively conversation. Delicate bouquets of exotic flowers added a pop of color to the scene, as playful and intricate paintings inspired by Fragonard, Watteau, and Boucher adorned the walls. The grand atmosphere is set for an unforgettable evening.}
    
    \centering
    \subfigure[CFG level=5, EP-CFG OFF]{
        \includegraphics[width=0.48\textwidth]{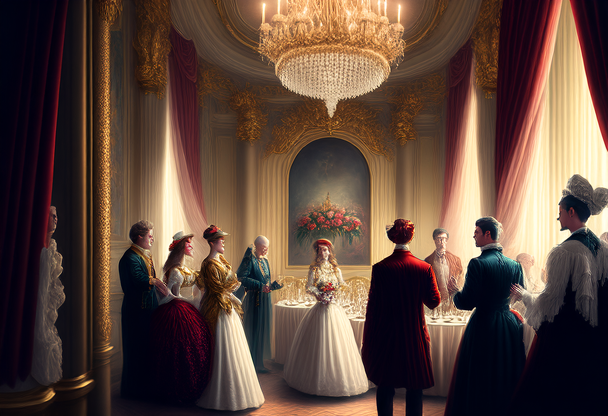}
    }
    \subfigure[CFG level=5, EP-CFG ON]{
        \includegraphics[width=0.48\textwidth]{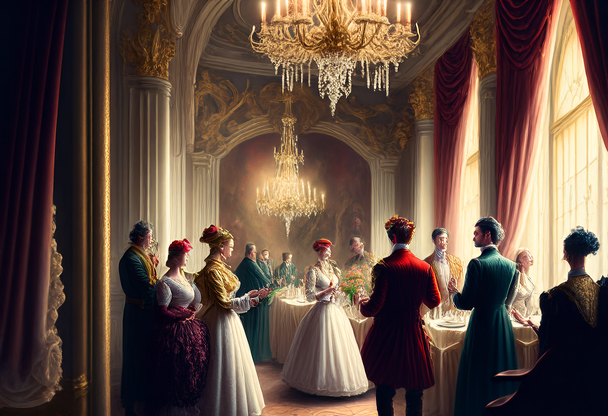}
    }
    
    \subfigure[CFG level=7, EP-CFG OFF]{
        \includegraphics[width=0.48\textwidth]{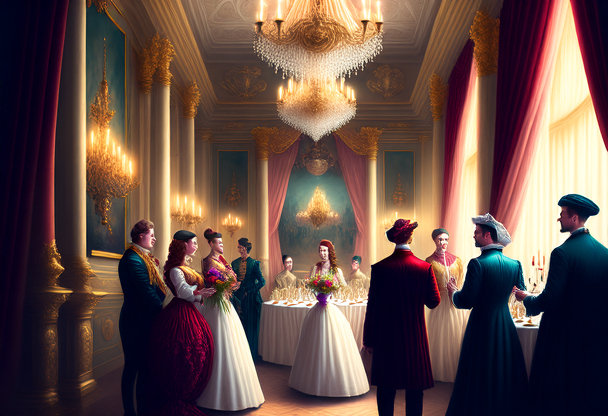}
    }
    \subfigure[CFG level=7, EP-CFG ON]{
        \includegraphics[width=0.48\textwidth]{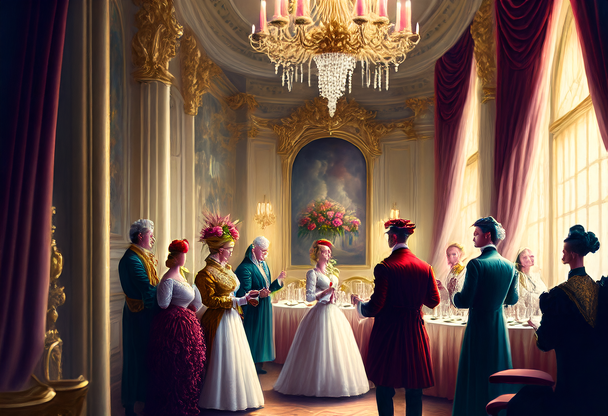}
    }
    
    \subfigure[CFG level=9, EP-CFG OFF]{
        \includegraphics[width=0.48\textwidth]{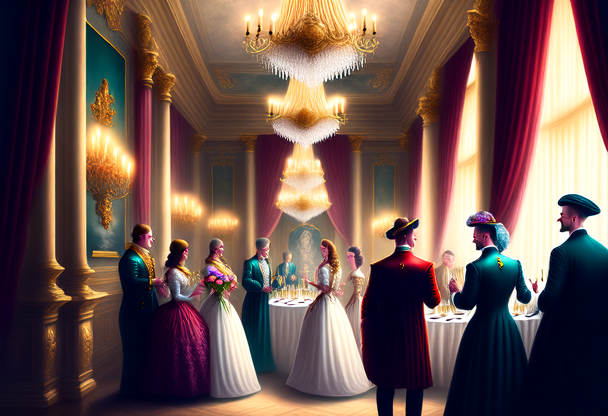}
    }
    \subfigure[CFG level=9, EP-CFG ON]{
        \includegraphics[width=0.48\textwidth]{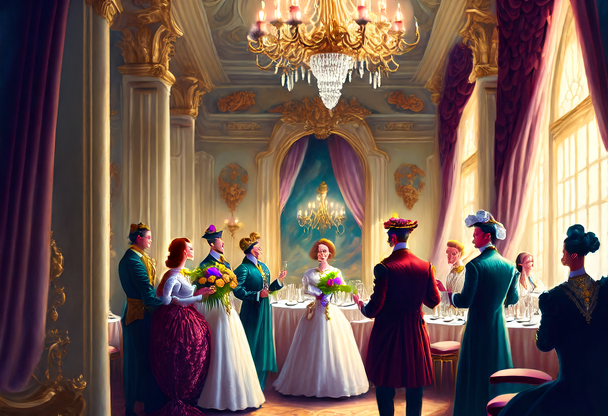}
    }
    
    \subfigure[CFG level=12, EP-CFG OFF]{
        \includegraphics[width=0.48\textwidth]{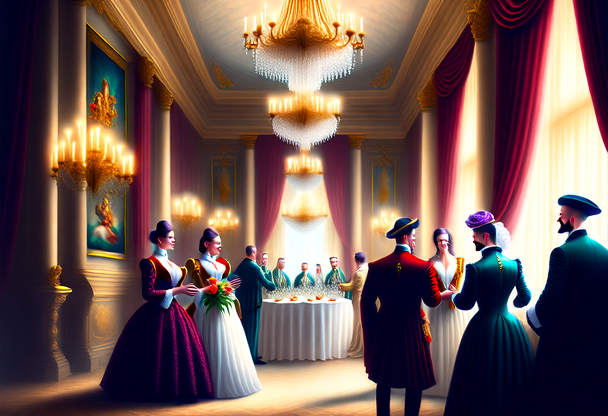}
    }
    \subfigure[CFG level=12, EP-CFG ON]{
        \includegraphics[width=0.48\textwidth]{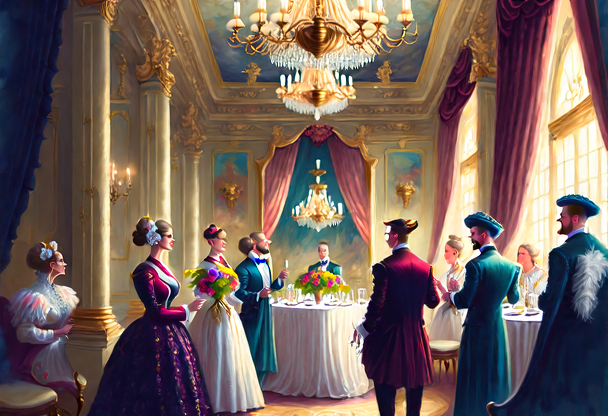}
    }
   
    \caption{Comparison of generation results with different CFG strengths (increasing from top to bottom). Without EP-CFG (left column), higher CFG strengths lead to over-contrast and light bloom effects in the chandelier lighting and ornate architectural details. EP-CFG (right column) maintains natural image quality and preserves the intricate period costumes and delicate banquet hall features across all CFG strengths while retaining the elegant Baroque ambiance.}
    \label{fig:filix_cfg_comparison7}
    \end{figurebox}
\end{figure}

\begin{figure}
\vspace*{-1.6cm}
\begin{figurebox}
    \textbf{Prompt:} \textit{Medium shot side profile portrait photo of the Takeshi Kaneshiro warrior chief, tribal panther make up, blue on red, looking away, serious eyes, 50mm portrait, photography, hard rim lighting photography.}
    
    \centering
    \subfigure[CFG level=5, EP-CFG OFF]{
        \includegraphics[width=0.48\textwidth]{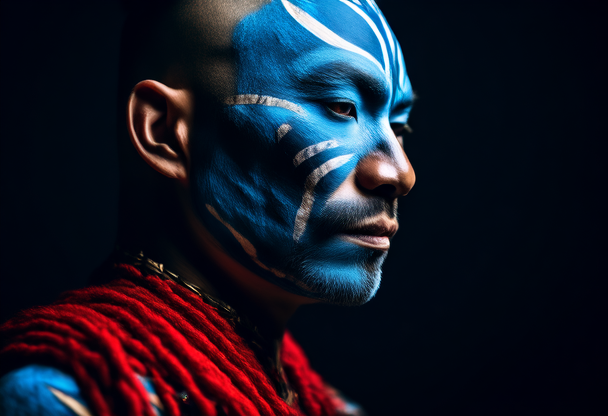}
    }
    \subfigure[CFG level=5, EP-CFG ON]{
        \includegraphics[width=0.48\textwidth]{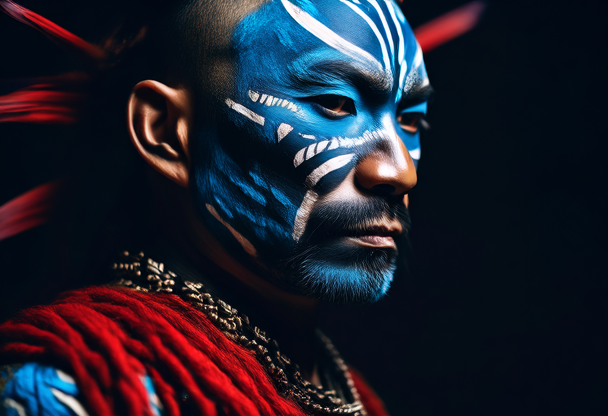}
    }
    
    \subfigure[CFG level=7, EP-CFG OFF]{
        \includegraphics[width=0.48\textwidth]{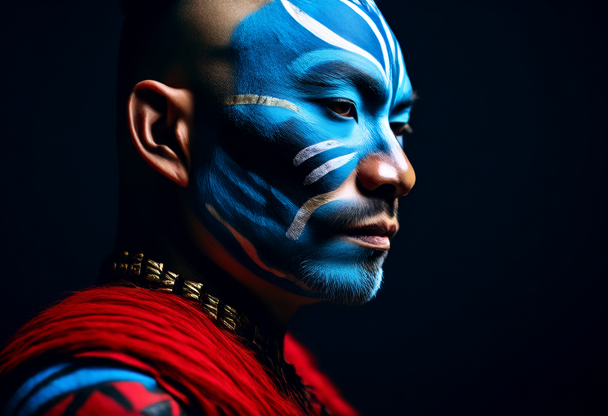}
    }
    \subfigure[CFG level=7, EP-CFG ON]{
        \includegraphics[width=0.48\textwidth]{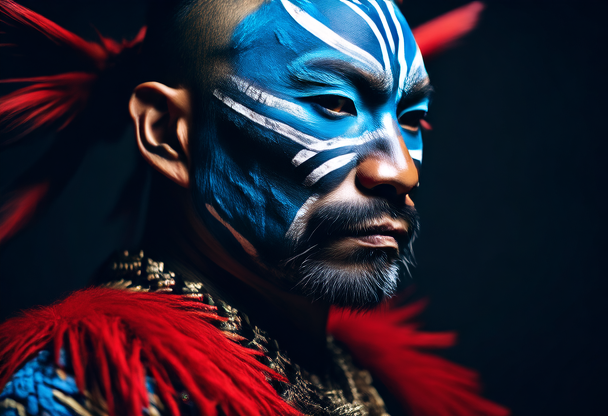}
    }
    
    \subfigure[CFG level=9, EP-CFG OFF]{
        \includegraphics[width=0.48\textwidth]{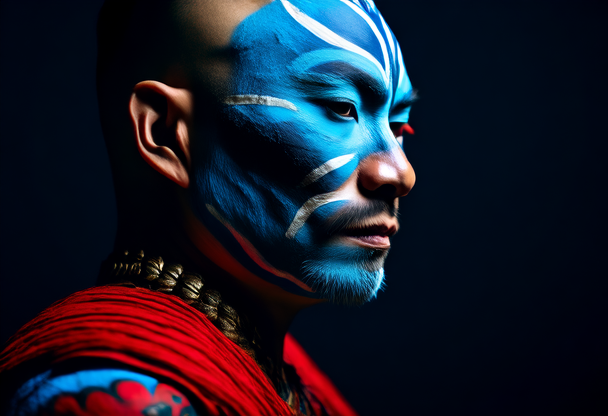}
    }
    \subfigure[CFG level=9, EP-CFG ON]{
        \includegraphics[width=0.48\textwidth]{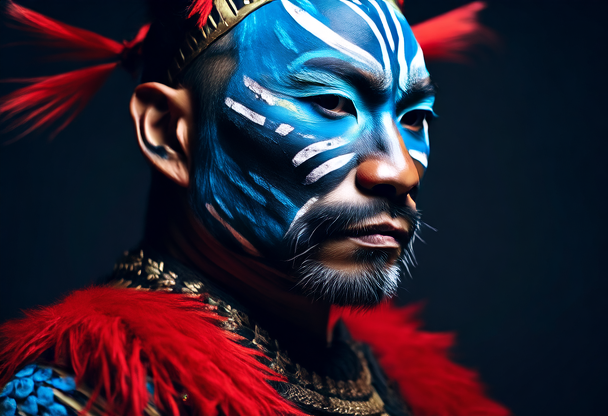}
    }
    
    \subfigure[CFG level=12, EP-CFG OFF]{
        \includegraphics[width=0.48\textwidth]{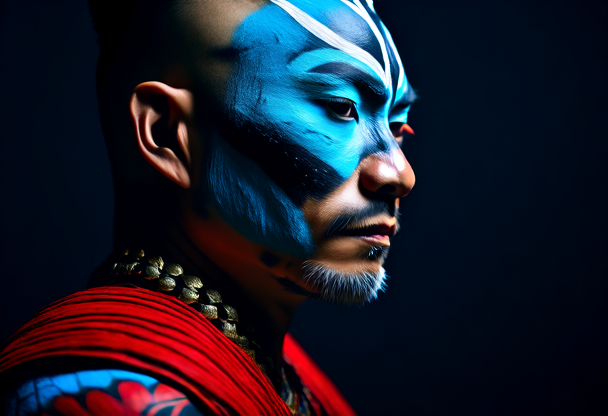}
    }
    \subfigure[CFG level=12, EP-CFG ON]{
        \includegraphics[width=0.48\textwidth]{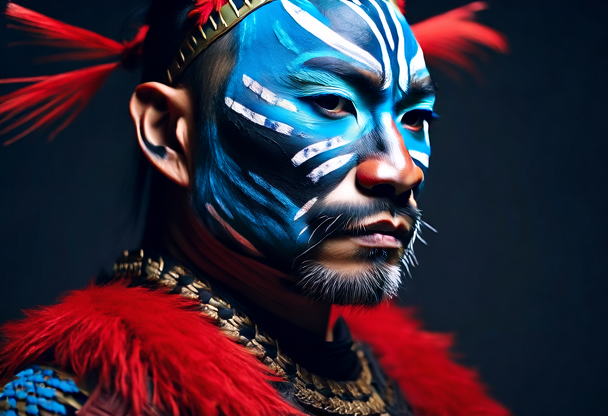}
    }
   
   \caption{Comparison of generation results with different CFG strengths (increasing from top to bottom). Without EP-CFG (left column), higher CFG strengths lead to over-contrast and loss of detail in the facial makeup patterns, while failing to maintain consistency in the ornate headpiece and feathered accessories. EP-CFG (right column) maintains natural image quality and preserves the intricate tribal designs, decorative headpiece, and textural details across all CFG strengths while retaining the dramatic portrait lighting.}
    \label{fig:filix_cfg_comparison8}
    \end{figurebox}
\end{figure}

\begin{figure}
\vspace*{-1.6cm}
\begin{figurebox}
    \textbf{Prompt:} \textit{A marble statue of the muse of history Clio with 'Adobe' written at the base.}
    
    \centering
    \subfigure[CFG level=5, EP-CFG OFF]{
        \includegraphics[width=0.48\textwidth]{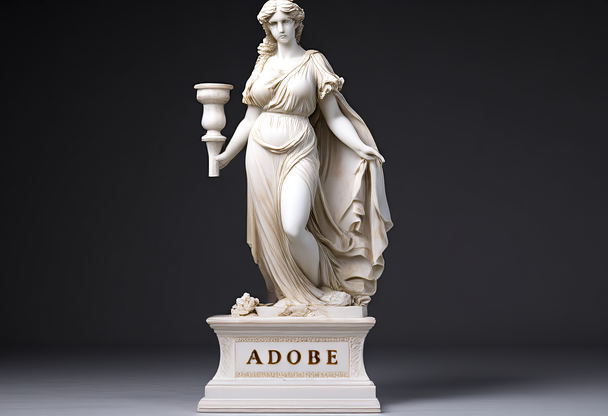}
    }
    \subfigure[CFG level=5, EP-CFG ON]{
        \includegraphics[width=0.48\textwidth]{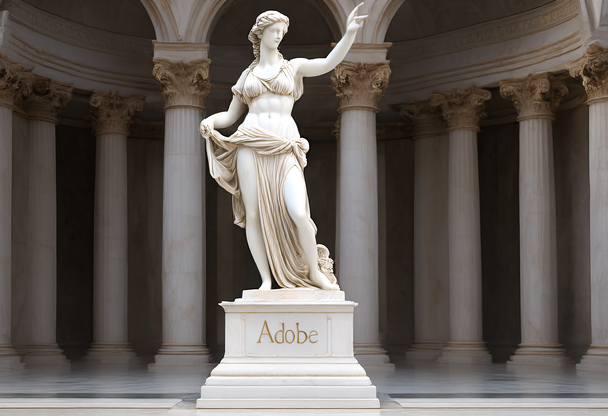}
    }
    
    \subfigure[CFG level=7, EP-CFG OFF]{
        \includegraphics[width=0.48\textwidth]{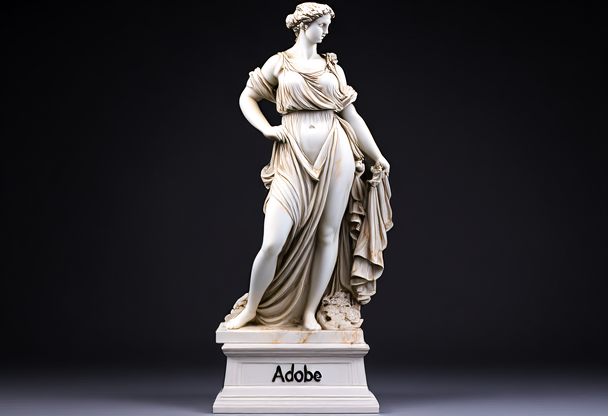}
    }
    \subfigure[CFG level=7, EP-CFG ON]{
        \includegraphics[width=0.48\textwidth]{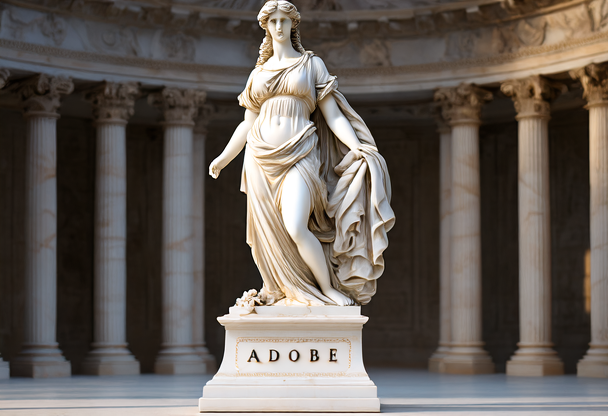}
    }
    
    \subfigure[CFG level=9, EP-CFG OFF]{
        \includegraphics[width=0.48\textwidth]{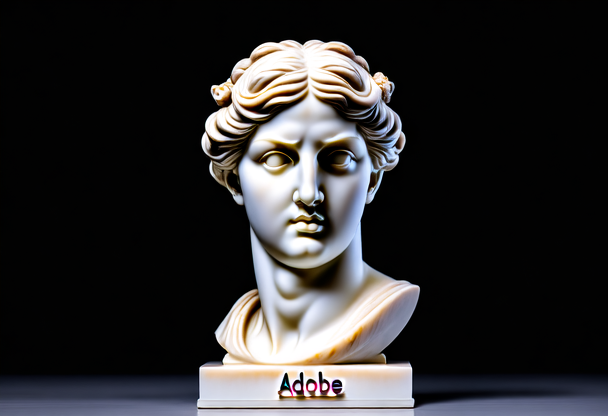}
    }
    \subfigure[CFG level=9, EP-CFG ON]{
        \includegraphics[width=0.48\textwidth]{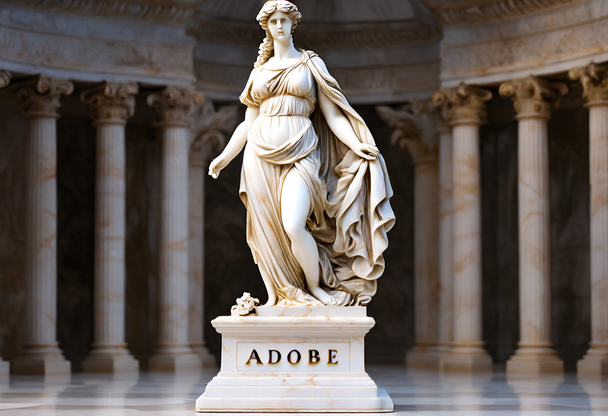}
    }
    
    \subfigure[CFG level=12, EP-CFG OFF]{
        \includegraphics[width=0.48\textwidth]{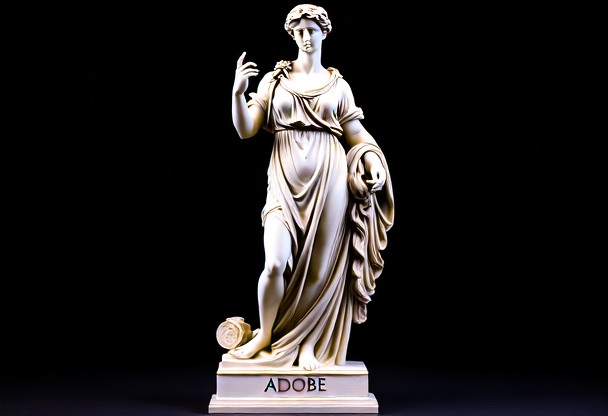}
    }
    \subfigure[CFG level=12, EP-CFG ON]{
        \includegraphics[width=0.48\textwidth]{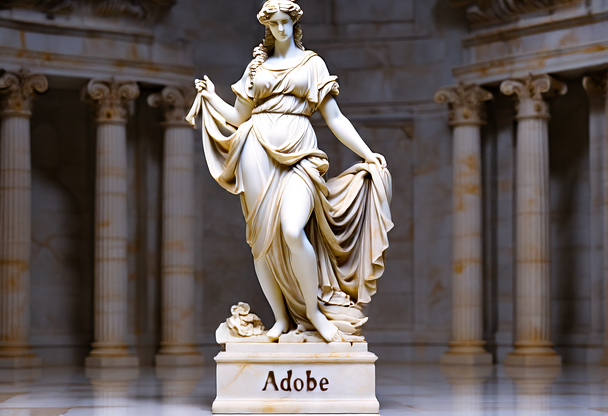}
    }
   
    \caption{Comparison of generation results with different CFG strengths (increasing from top to bottom). Without EP-CFG (left column), higher CFG strengths lead to over-contrast and inconsistent rendering of the marble texture and statue proportions. EP-CFG (right column) maintains natural image quality and preserves the classical sculptural details across all CFG strengths while retaining the architectural context of the columned setting.}
    \label{fig:filix_cfg_comparison9}
    \end{figurebox}
\end{figure}

\begin{figure}
\vspace*{-1.6cm}
\begin{figurebox}
    \textbf{Prompt:} \textit{An IT-guy trying to fix hardware of a PC tower is being tangled by the PC cables like Laokoon. Marble, copy after Hellenistic original from ca. 200 BC. Found in the Baths of Trajan, 1506.}
    
    \centering
    \subfigure[CFG level=5, EP-CFG OFF]{
        \includegraphics[width=0.48\textwidth]{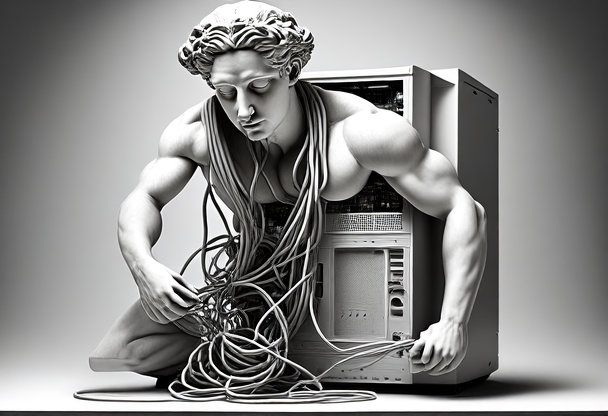}
    }
    \subfigure[CFG level=5, EP-CFG ON]{
        \includegraphics[width=0.48\textwidth]{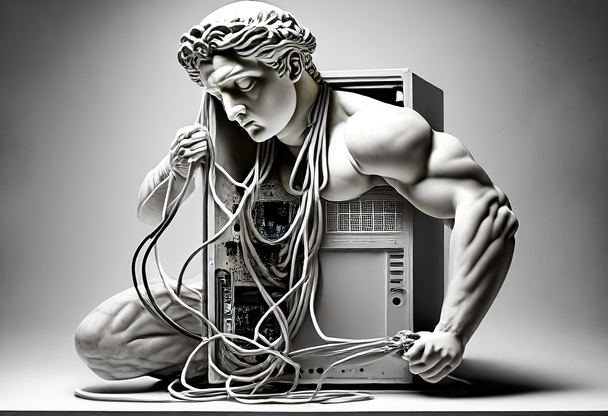}
    }
    
    \subfigure[CFG level=7, EP-CFG OFF]{
        \includegraphics[width=0.48\textwidth]{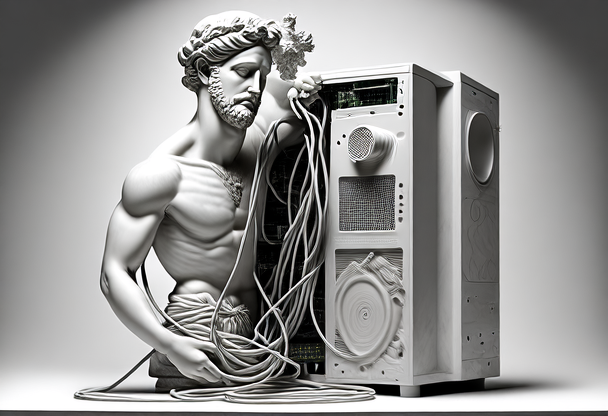}
    }
    \subfigure[CFG level=7, EP-CFG ON]{
        \includegraphics[width=0.48\textwidth]{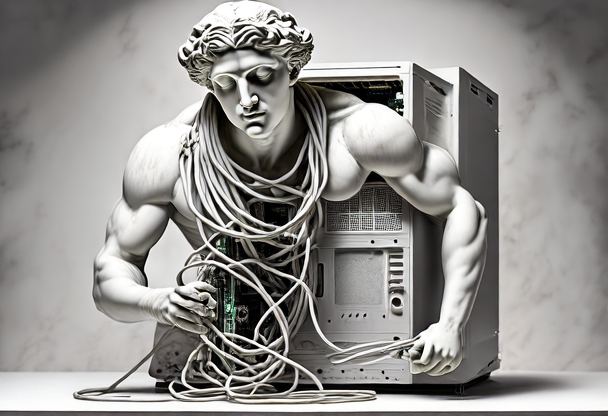}
    }
    
    \subfigure[CFG level=9, EP-CFG OFF]{
        \includegraphics[width=0.48\textwidth]{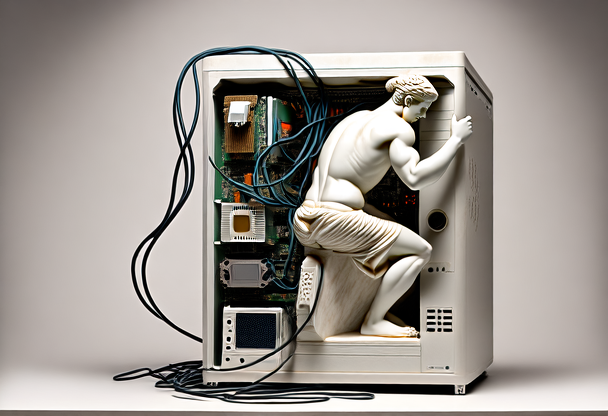}
    }
    \subfigure[CFG level=9, EP-CFG ON]{
        \includegraphics[width=0.48\textwidth]{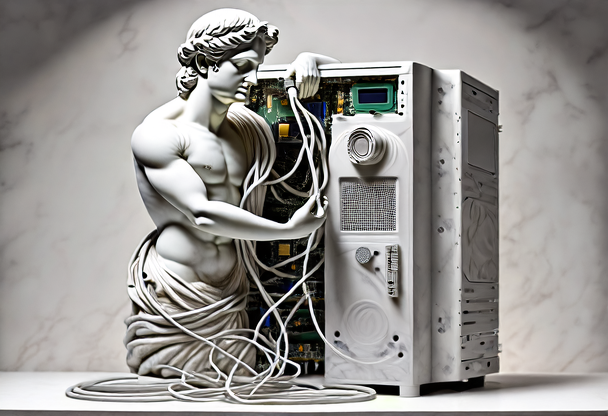}
    }
    
    \subfigure[CFG level=12, EP-CFG OFF]{
        \includegraphics[width=0.48\textwidth]{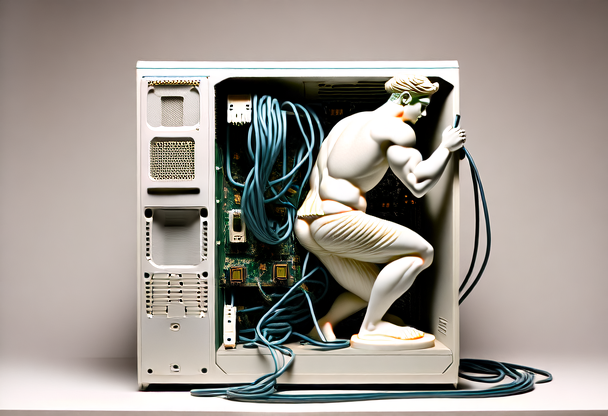}
    }
    \subfigure[CFG level=12, EP-CFG ON]{
        \includegraphics[width=0.48\textwidth]{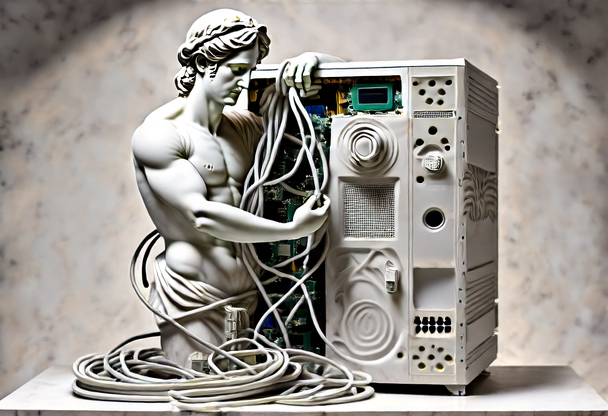}
    }
   
    \caption{Comparison of generation results with different CFG strengths (increasing from top to bottom). Without EP-CFG (left column), higher CFG strengths lead to inconsistent rendering of the classical sculpture style and PC components, with varying compositions at each level. EP-CFG (right column) maintains consistent sculptural quality and preserves both the Hellenistic aesthetics and modern technological elements across all CFG strengths while retaining the satirical fusion of ancient and contemporary themes.}
    \label{fig:filix_cfg_comparison10}
    \end{figurebox}
\end{figure}

\begin{figure}
\vspace*{-1.6cm}
\begin{figurebox}
    \textbf{Prompt:} \textit{Photo of a mega Lego space station inside a kid's bedroom.}
    
    \centering
    \subfigure[CFG level=5, EP-CFG OFF]{
        \includegraphics[width=0.48\textwidth]{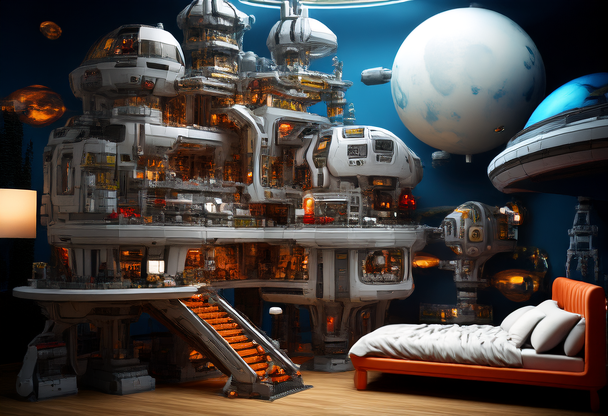}
    }
    \subfigure[CFG level=5, EP-CFG ON]{
        \includegraphics[width=0.48\textwidth]{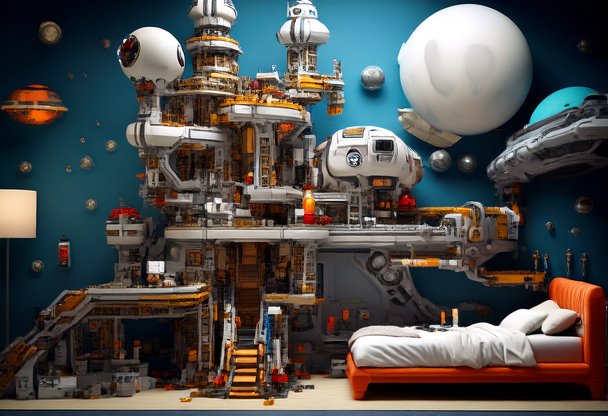}
    }
    
    \subfigure[CFG level=7, EP-CFG OFF]{
        \includegraphics[width=0.48\textwidth]{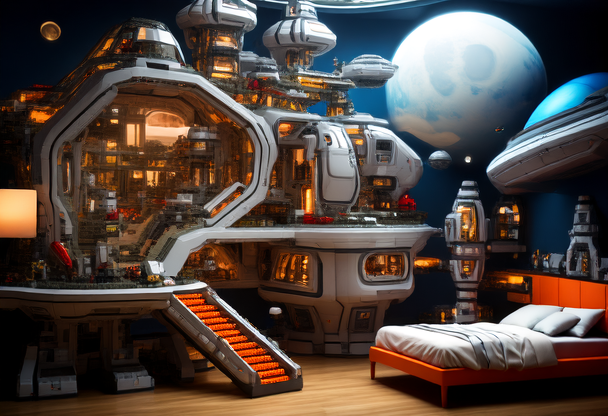}
    }
    \subfigure[CFG level=7, EP-CFG ON]{
        \includegraphics[width=0.48\textwidth]{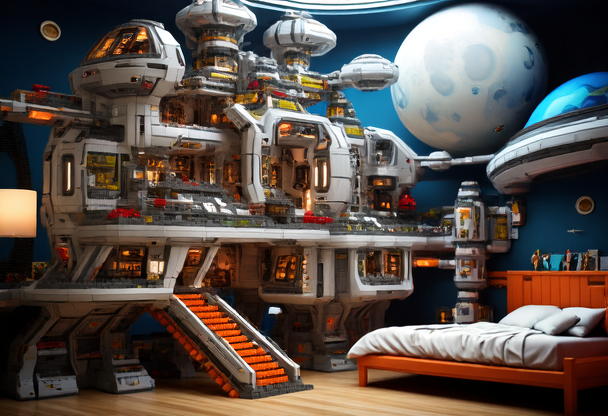}
    }
    
    \subfigure[CFG level=9, EP-CFG OFF]{
        \includegraphics[width=0.48\textwidth]{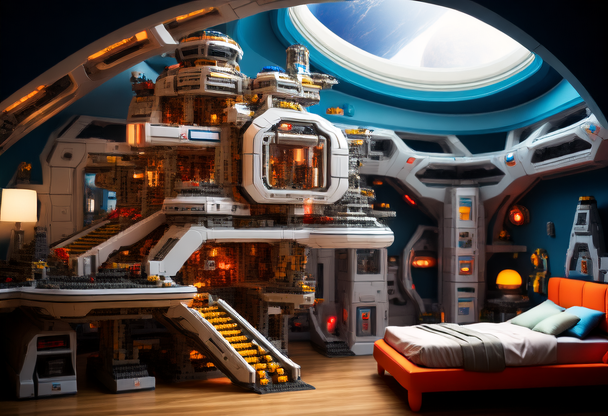}
    }
    \subfigure[CFG level=9, EP-CFG ON]{
        \includegraphics[width=0.48\textwidth]{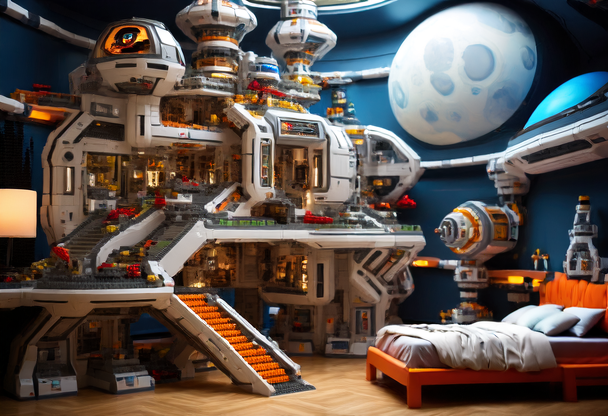}
    }
    
    \subfigure[CFG level=12, EP-CFG OFF]{
        \includegraphics[width=0.48\textwidth]{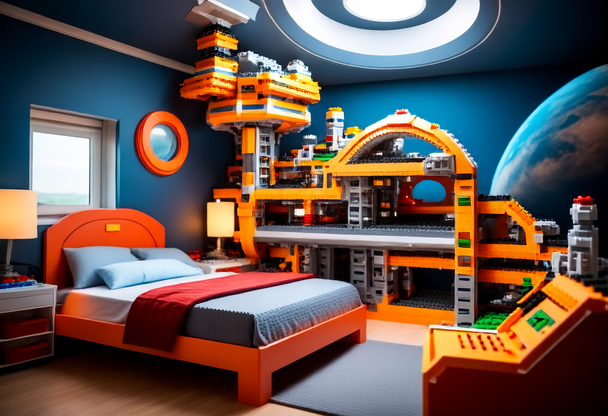}
    }
    \subfigure[CFG level=12, EP-CFG ON]{
        \includegraphics[width=0.48\textwidth]{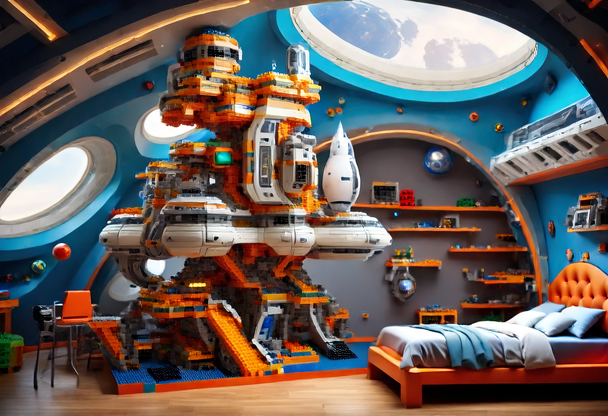}
    }
   
     \caption{Comparison of generation results with different CFG strengths (increasing from top to bottom). Without EP-CFG (left column), higher CFG strengths lead to inconsistent architecture and drastic variations in the space station's design, eventually losing the Lego aesthetic and the sci-fi space station style at the highest level. EP-CFG (right column) maintains consistent structural details and preserves the intricate Lego construction elements across all CFG strengths while retaining the imaginative bedroom setting.}
    \label{fig:filix_cfg_comparison11}
    \end{figurebox}
\end{figure}

\begin{figure}
\vspace*{-1.6cm}
\begin{figurebox}
    \textbf{Prompt:} \textit{Starting from a ground-level view of a road leading towards a tunnel in a graffiti covered city with skyscrapers in the background, the camera tracks smoothly along the road into a short dark tunnel. As it emerges on the other side, the camera rapidly ascends, revealing the road continuing through a huge field of multicoloured wildflowers surrounded by snow capped mountains.}
    
    \centering
    \subfigure[CFG level=5, EP-CFG OFF]{
        \includegraphics[width=0.48\textwidth]{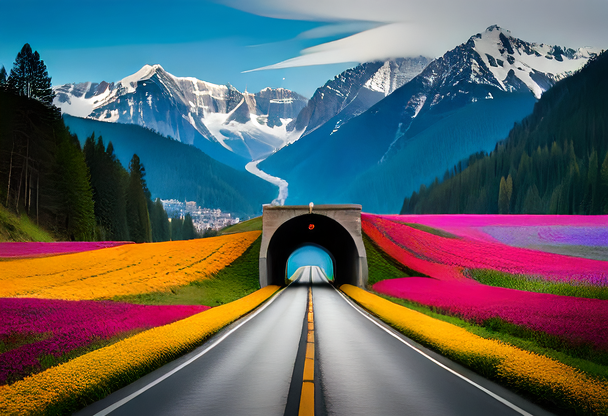}
    }
    \subfigure[CFG level=5, EP-CFG ON]{
        \includegraphics[width=0.48\textwidth]{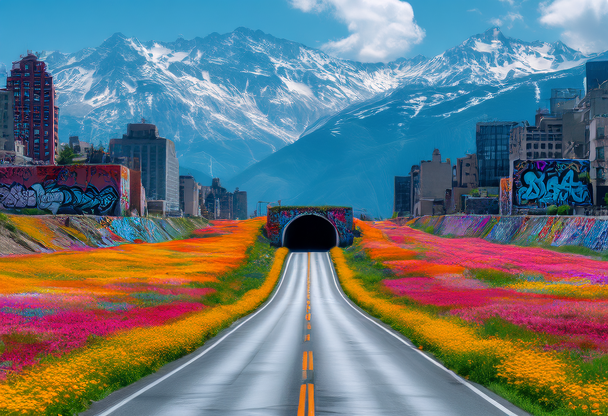}
    }
    
    \subfigure[CFG level=7, EP-CFG OFF]{
        \includegraphics[width=0.48\textwidth]{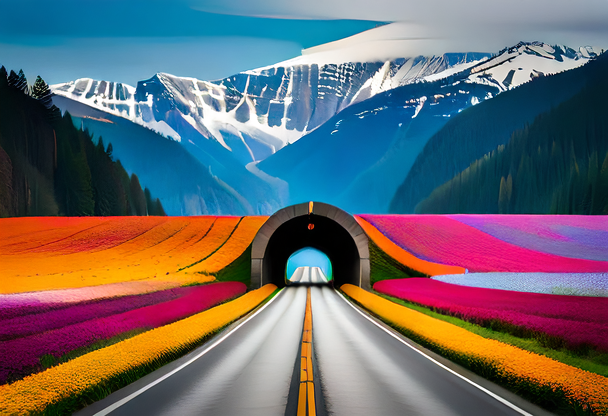}
    }
    \subfigure[CFG level=7, EP-CFG ON]{
        \includegraphics[width=0.48\textwidth]{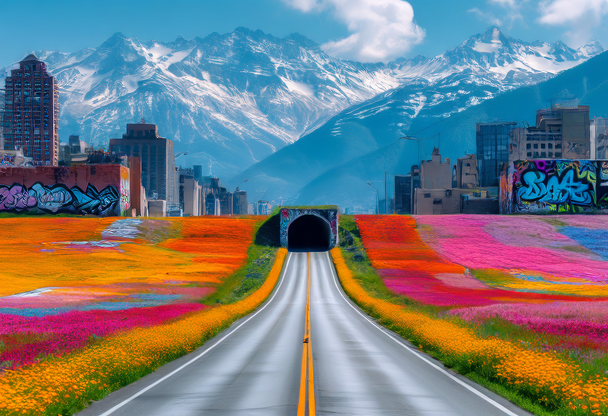}
    }
    
    \subfigure[CFG level=9, EP-CFG OFF]{
        \includegraphics[width=0.48\textwidth]{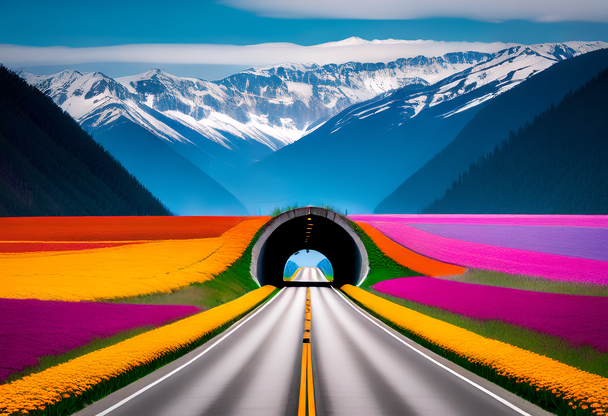}
    }
    \subfigure[CFG level=9, EP-CFG ON]{
        \includegraphics[width=0.48\textwidth]{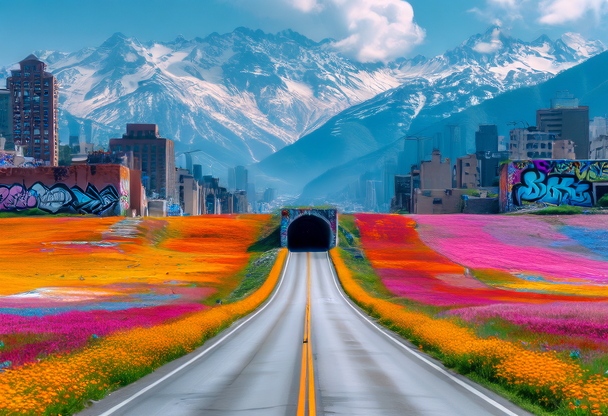}
    }
    
    \subfigure[CFG level=12, EP-CFG OFF]{
        \includegraphics[width=0.48\textwidth]{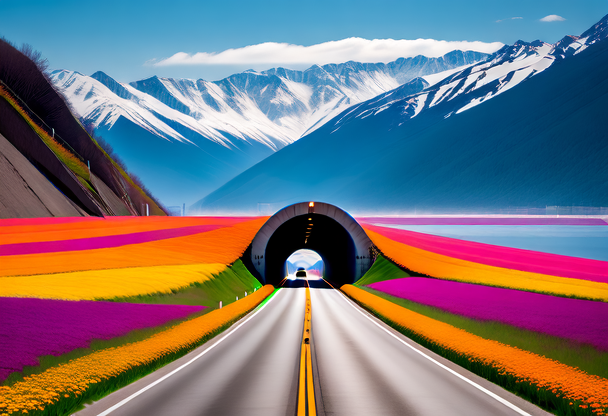}
    }
    \subfigure[CFG level=12, EP-CFG ON]{
        \includegraphics[width=0.48\textwidth]{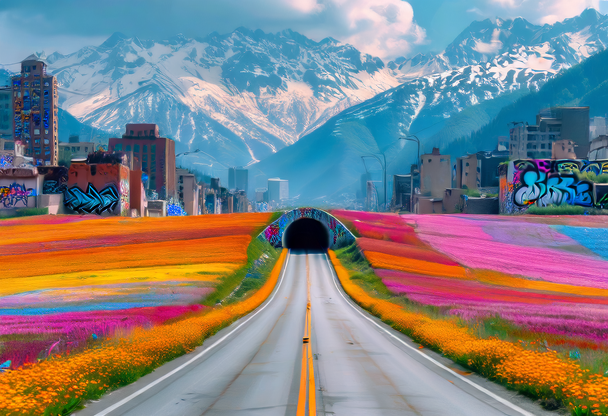}
    }
   
    \caption{Comparison of generation results with different CFG strengths (increasing from top to bottom). Without EP-CFG (left column), higher CFG strengths lead to over-contrast in the mountain scenery and missing urban elements entirely, focusing solely on the nature view. EP-CFG (right column) maintains consistent balance between the urban and natural environments, preserving both the graffiti-covered buildings and wildflower fields across all CFG strengths while retaining the dramatic transition composition.}
    \label{fig:filix_cfg_comparison12}
    \end{figurebox}
\end{figure}